\documentclass{article}

\usepackage[preprint]{corl_2026} 
\usepackage{graphics} 
\usepackage{amsmath} 
\usepackage{epsfig} 
\usepackage{times} 
\usepackage{amsmath} 
\usepackage{amssymb}  
\usepackage[dvipsnames]{xcolor}
\usepackage{booktabs}
\usepackage{makecell}
\usepackage{pifont} 
\usepackage{colortbl}
\usepackage{amssymb}
\usepackage{enumitem}
\usepackage{booktabs}
\usepackage{colortbl}
\usepackage{xcolor}
\usepackage{makecell}
\usepackage{tabularx} 
\usepackage{multirow} 
\usepackage{wrapfig}

\definecolor{fixedgray}{gray}{0.93}
\definecolor{darkgreen}{RGB}{59,125,35}
\definecolor{light_purple}{RGB}{216,110,204}
\definecolor{dark_blue}{RGB}{8, 79, 106}
\definecolor{dark_purple}{RGB}{102,0,102}
\definecolor{Grey}{rgb}{0.5, 0.5, 0.5}

\definecolor{pillbg}{HTML}{B5D4F4}
\definecolor{rowblue}{HTML}{E6F1FB}
\definecolor{ourstext}{HTML}{185FA5}
\usepackage{amssymb}
\usepackage{xcolor}
\usepackage{adjustbox}

\usepackage{bm}
\newcommand{\cmark}{{\color{teal}\rlap{$\checkmark$}$\checkmark$}}
\newcommand{\xmark}{{\color{red!70}\rlap{$\boldsymbol{\times}$}$\boldsymbol{\times}$}}
\usepackage{silence}
\title{From Region Arrival to Instance-Level Grounding in Vision-and-Language Navigation}

\author{
  Xiangyu Shi$^{1}$, Ruoxi Yang$^{2}$, Wei Tao$^{2}$, Jiwen Zhang$^{3}$, Yanyuan Qiao$^{\dagger{4}}$,  Qi Wu $^{\ddagger1}$
  \\
  $^{1}$Australian Institute for Machine Learning, Adelaide University \\
  $^{2}$Ruyi Dynamics\\$^{3}$Fudan University\\ $^{4}$Swiss Federal Institute of
Technology Lausanne (EPFL)\\
  \\
  $^\dagger$ Project Lead, 
  $^\ddagger$ Corresponding Author\\
}

\begin{document}
\maketitle


\begin{abstract}
Vision-and-Language Navigation (VLN) agents may satisfy conventional success criteria while still failing to establish reliable object-level grounding, because current evaluation protocols mainly reward stopping within a 3-meter radius and largely ignore the agent’s final orientation and target visibility. We formalize this limitation as the Last-3-Meter Grounding Gap and introduce three instance-centric metrics to quantify proximity precision, target visibility, and final-view grounding. To mitigate this gap, we propose REALM (Region-to-Entity Alignment for Last-3-Meter Navigation), a plug-and-play, architecture-agnostic refinement module that decouples fine-grained target approaching from long-horizon navigation. REALM uses a visibility-aware stopping strategy to reduce premature termination and improve final viewpoint alignment. We further construct REVERIE-AIM, which provides object-instance-level goals and 180K short-horizon training samples for final-stage target approaching. Extensive evaluations across four diverse VLN backbones show that REALM consistently improves proximity precision and visual grounding success, demonstrating its broad applicability.
\end{abstract}

\keywords{Vision-and-Language Navigation, Object-Centric Navigation} 

\section{Introduction}

For embodied agents to follow language instructions in physical environments, reaching the correct room or region is often insufficient: the agent must also obtain a viewpoint from which the referred object can be reliably perceived and grounded. This capability bridges high-level language understanding with object-level visual grounding, and provides a necessary perceptual interface for downstream interaction tasks such as inspection or manipulation. The REVERIE task~\cite{qi2020reverie} was the first introduced to study this setting: given a natural language instruction containing a referring expression, the agent must navigate through an indoor environment, locate the specified object instance, and predict its bounding box. REVERIE-CE~\cite{reve-ce} further transfers this task from discrete navigation graphs to continuous environments, bringing it closer to realistic robotic deployment.

\begin{figure}[t]
\setlength{\abovecaptionskip}{-2pt}
    \centering
    \includegraphics[width=\linewidth]{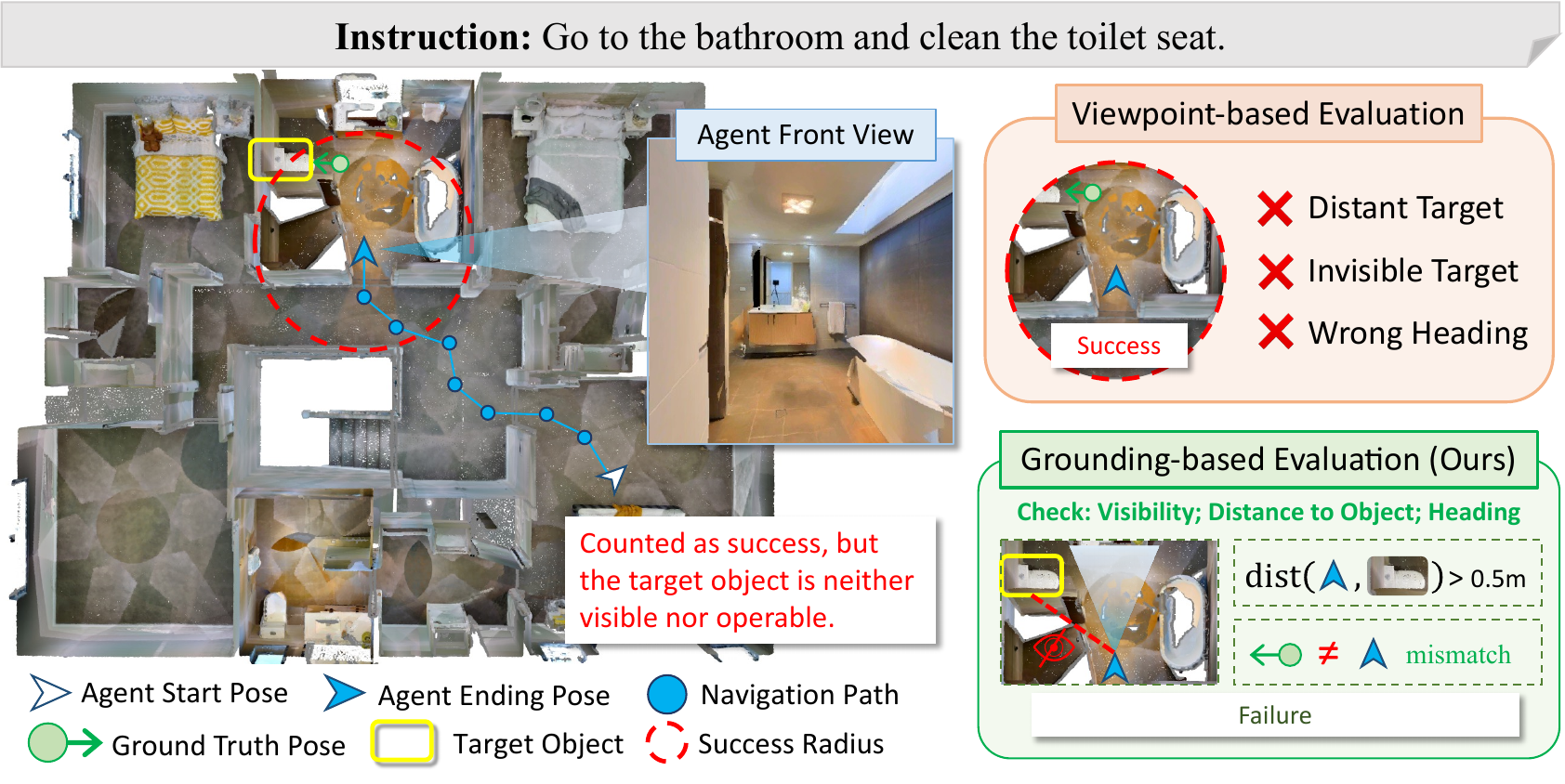}
    \caption{Existing evaluation rewards stopping near a target viewpoint, but object grounding requires a final viewpoint where the referred object is actually visible and localizable.}
    \label{fig:intro}
    \vspace{-19pt}
\end{figure}

However, current evaluation protocols for REVERIE and REVERIE-CE still largely inherit region-level navigation criteria from VLN-CE. In subsequent methods~\cite{dynam3d,d3d}, an agent is typically deemed successful if it stops within 3 meters of a pre-annotated endpoint. This endpoint represents a navigable location near the target object, rather than the object instance itself. Moreover, the agent's final heading is not evaluated: it may stop within a 3m radius while facing a wall, a corridor, or the opposite direction from the target. As a result, a trajectory can be counted as successful even when the referred object is not visible from the final viewpoint. This critical flaw is illustrated in Figure~\ref{fig:intro}: while existing evaluation protocols reward the agent for merely stopping near a pre-annotated target region, true object grounding requires the agent to actively adjust its final viewpoint so that the referred object is actually visible and localizable in its camera frame. For tasks that require object-level grounding, such as confirming the object's presence, inspecting its state, or preparing for manipulation, such region-level arrival does not provide a functional endpoint.

While related problems have been studied in other embodied navigation tasks, such as switching strategies near the target~\cite{sling}, optimizing viewpoints via point-cloud scoring~\cite{msgnav}, and recovering 6-DoF pose from a reference image~\cite{deng2026anyimagenav}, these methods assume successful target localization or a known goal image, neither of which applies to the VLN setting where the agent must ground a free-form language instruction. We term this systematic gap the \textbf{Last-3-Meter Grounding Gap}: the disconnect between coarse-grained region arrival and fine-grained object grounding. Bridging this gap is essential for deploying language-conditioned navigation in any real robotic system where the agent must subsequently perceive, manipulate, or interact with the referred object.

To quantify this gap, we propose three complementary object-instance-centric metrics: Object Navigation Success (ONS) for geodesic proximity to the target instance, Grounding Success (GS) for detection accuracy at the final viewpoint, and Oracle Grounding  Success (OracleGS) for target visibility in the camera frame. Under these metrics, we find that even ETPNav-FT~\cite{an2024etpnav}, fine-tuned on the REVERIE-CE dataset, achieves only 6.32\% ONS@0.1m and 11.31\% OracleGS, in sharp contrast to its 34.67\%  Success Rate under the conventional 3-meter criterion. This gap shows that strong performance under the conventional Success Rate does not necessarily carry over to instance-level positioning and grounding: methods that rank well on standard VLN metrics can still perform poorly once the target must actually be reached and faced.

To bridge this gap, we then propose \textbf{REALM} (Region-to-Entity Alignment for Last-3-Meter Navigation), a plug-and-play refinement module decoupled from the upstream navigation policy. This decoupled design is motivated by two considerations. Firstly, long-horizon navigation and short-horizon target approach are fundamentally different problems: the former demands path planning and scene-level reasoning over tens of meters, while the latter requires precise visual alignment and visibility-grounded stopping within a few meters. Coupling them into a single model forces them to share capacity, yielding suboptimal fine-grained behaviour which limits fine-grained performance even when the upstream policy is strong. Secondly, the VLN landscape is highly diverse and rapidly evolving, with methods spanning graph-based, video-based, zero-shot, and training-free paradigms. Tying fine-grained positioning to any specific architecture would require rebuilding the alignment module whenever the upstream model changes, which is neither generalizable nor sustainable. Once any VLN model issues a stop command, REALM takes over using only egocentric RGB and the original language instruction, with no depth sensors, maps, or architectural changes to the upstream model. We instantiate REALM by LoRA-finetuning UniNaVid~\cite{uninavid} with a tailored visibility-aware stopping strategy, which successfully suppresses premature termination and triggers decisive stops based on explicit object perception. In addition, to support training and evaluation, we construct \textbf{REVERIE-AIM}, which replaces region-level endpoints with object-instance-level goals and provides 180K short-horizon samples for the final approach stage.

Extensive evaluations across four diverse VLN backbones demonstrate that REALM consistently and significantly bridges the Last-3-Meter Grounding Gap. For instance, it elevates ETPNav-ZS's ONS@0.1m from 7.07\% to 14.66\% (a 107\% relative gain) and substantially boosts SmartWay's OracleGS from 4.69\% to 15.92\%. Ablation results confirm that our complete loss function brings consistent improvements by successfully suppressing premature termination. Furthermore, physical deployment on a Hello Robot Stretch platform provides preliminary evidence of real-world applicability, escalating the real-world navigation Success Rate (ONS@0.5m) from 8.33\% to 33.33\%.

To summarise, our contributions are: (1) we identify the \textbf{Last-3-Meter Grounding Gap}: the disconnect between reaching the correct region and actually perceiving the target object; (2) we propose  \textbf{REALM}, a plug-and-play decoupled refinement module with a visibility-aware stopping strategy; (3) we construct the  \textbf{REVERIE-AIM} dataset and three evaluation metrics (ONS, GS, OracleGS); and (4) we validate REALM across four diverse VLN backbones, demonstrating consistent improvements in fine-grained grounding.

\vspace{-5pt}
\section{Related Work}

\vspace{-5pt}
Vision-and-Language Navigation (VLN) requires an autonomous agent to navigate by following natural language instructions~\cite{anderson2018vision,zhang2024vision,su2026capnav,wang2026progress,jin2026does}. 
Early benchmarks primarily operate on discrete environments, where the agent’s movement is restricted to predefined navigation graphs~\cite{qi2020reverie,anderson2020rxr,thomason2020cvdn,qiao2022hop,zhang2021curriculum}. 
To better align with real-world applications, Krantz et al.~\cite{krantz2020beyond} lifted these constraints by extending VLN to continuous environments (VLN-CE), enabling agents to move freely without a predefined topological map. 

Beyond standard instruction-following VLN, REVERIE~\cite{qi2020reverie} introduces a more complex setting that requires agents to navigate via high-level instructions and localise a specific target object. 
REVERIE-CE~\cite{reve-ce} further transitions this task into VLN-CE, bringing low-level control and obstacle avoidance challenges into focus. To tackle this, specialised methods like Dynam3D~\cite{dynam3d} and D3D-VLP~\cite{d3d} leverage dynamic 3D representations on REVERIE-CE, empowering vision-language models with spatial tokens or dynamic chain-of-thought pipelines. Alternatively, advanced paradigms from broader VLN-CE domains, such as graph-based models~\cite{an2024etpnav,zhang2025cosmo}, video-based models~\cite{uninavid}, or zero-shot navigation approaches~\cite{InstructNav,smartway,chen2025constraint}, can also be adapted to this task. Concurrently, GroundingMate~\cite{groundingmate} identifies the gap between navigation success and object grounding success in discrete REVERIE, leveraging an MLLM to assist target selection when the agent is uncertain. While effective, its object grounding remains a classification-based paradigm that selects from pre-annotated candidate bounding boxes rather than performing open-world detection, and it operates exclusively in discrete graph environments. However, we observe that both specialised and broader continuous methods inherit a legacy evaluation protocol: success depends solely on stopping within 3 meters of an endpoint, largely ignoring orientation accuracy or object grounding precision. Our work identifies and systematically addresses this critical evaluation gap.

Beyond coarse trajectory following, several works have begun to address fine-grained target approaching. 
SLING~\cite{sling} switches navigation strategies upon nearing the target. MSGNav~\cite{msgnav} optimises the 
observation viewpoint via point-cloud visibility scoring after target localisation. AnyImageNav~\cite{deng2026anyimagenav} 
extends the InstanceImageNav success criterion from coarse distance thresholds to precise 6-DoF pose recovery 
by geometrically registering the goal-image camera orientation. More recently, APRR~\cite{qin2025active} decouples navigation into an exploration phase and a ``last-mile" phase. The latter dynamically refines the robot's pose via active perception to ensure an optimal viewpoint for manipulation. However, these methods either assume successful target localisation as a precondition or rely on a known reference image as the navigation goal, neither of which directly transfers to the VLN setting. More importantly, none of them considers the VLN evaluation framework itself: current benchmarks judge success solely by endpoint proximity while imposing no constraint on final orientation, leaving the \textit{Last-3-Meter Grounding Gap} entirely unaddressed.

\section{The Last-3-Meter Grounding Problem}
\subsection{Preliminaries}
\label{sec:VLN-CE-Task}

For the REVERIE-CE~\cite{reve-ce} task, at each step $t$, the navigation policy $\pi_{\text{nav}}$ receives an observation $\mathcal{O}_t$, which may consist of various sensory inputs depending on the system design, including egocentric or panoramic RGB-D views, RGB-only inputs, or other modality-specific features. The agent state at time $t$ is defined as $s_t = (p_t, \theta_t)$, where $p_t \in \mathbb{R}^3$ denotes the 3D position and $\theta_t$ denotes the heading direction.
Given a high-level natural language instruction $\mathcal{I}$, the agent is required to navigate in an unseen indoor environment $\mathcal{E}$ and reach the goal location.
After $\pi_{\text{nav}}$ issues a \texttt{stop} command at step ${T_{\text{nav}}}$, the agent needs localise the referred target object $o^*$ within its final observation $\mathcal{O}_{T_{\text{nav}}}$ by predicting a bounding box $\hat{b}$:
$\hat{b} = \mathrm{Det}(\mathcal{O}_{T_{\text{nav}}},\, \mathcal{I})$

\subsection{Problem Formulation: The Last-3-Meter Grounding Gap}

In standard VLN-CE evaluation, an episode is deemed successful if the agent's stopping position $p_{T_{\text{nav}}}$ lies within 3\,meters of a pre-annotated endpoint $p_{\text{ep}}$:
$\text{Success Rate (SR)} = \left[ d(p_{T_{\text{nav}}},\, p_{\text{ep}}) \leq 3\,\text{m} \right].$

However, this paradigm fails in instance-level language instructions~\cite{qi2020reverie,reve-ce} because $p_{\text{ep}}$ is merely a neighbouring waypoint rather than the physical target. For practical embodied instruction following, the final state $s_{T_{nav}} = (p_{T_{\text{nav}}}, \theta_{T_{\text{nav}}})$ must satisfy two conditions: (i)~\textbf{Physical proximity} to the actual target instance, and (ii)~\textbf{Visual groundability} in the egocentric view with sufficient resolution.

These two conditions are purely task-level requirements independent of any particular model design, and together define what we term the \textbf{last-3-meter grounding gap}: the discrepancy between reaching the general vicinity of the target (as measured by SR) and achieving a final state from which reliable instance-level grounding is possible.

\subsection{Evaluation Metrics}
\label{sec:metrics}
Prior evaluation protocols typically assess navigation and grounding jointly. In contrast, we decouple them into three independent axes to enable fine-grained diagnosis.

\vspace{-8pt}
\paragraph{Object Navigation Success (ONS).}
Following ObjectNav~\cite{hssd}, we measure navigation precision 
via the minimum geodesic distance from the agent's stop position 
to navigable points surrounding the target:
\[
    \text{ONS}@\tau = \!\left[\min_{q \in \mathcal{S}_{\text{obj}}} 
    d(p_{T_{\text{nav}}},\, q) \leq \tau\right],
\]
where $\mathcal{S}_{\text{obj}}$ is the set of navigable points 
sampled around the target instance and $d(\cdot,\cdot)$ is geodesic 
distance. Unlike ObjectNav, we require proximity to the specified 
instance rather than any category match. We report $\tau{=}0.1$\,m 
(standard ObjectNav) and $\tau{=}0.5$\,m (for downstream positioning 
tolerance). Note that VLN-CE's SR measures alignment with the 
annotated endpoint, whereas ONS directly measures proximity to 
the target object.

\vspace{-8pt}
\paragraph{Grounding Success (GS).}
Grounding success is evaluated unconditionally, independent of whether navigation succeeded. Given the predicted bounding box $\hat{b}$ produced by the detection module at the agent's final viewpoint, and the ground-truth bounding box $b^{*}$, we define:
\[
    \text{GS} = \!\left[\text{IoU}(\hat{b},\, b^{*}) \geq 0.5\right].
\]

\vspace{-8pt}
\paragraph{Oracle Grounding Success (OracleGS).}
An episode achieves oracle grounding success if the ground-truth bounding box $b^{*}$ of the target instance is visible in the agent's current observation $\mathcal{O}_{T}$ and occupies at least 1\% of the camera frame area:
\[
    \text{OracleGS} = \left[\frac{\text{Area}(b^{*})}{\text{Area}(\mathcal{O}_{T})} \geq 0.01\right].
\]
By comparing GS with OracleGS, we can distinguish visibility failures caused by suboptimal navigation from failures caused by the detection module.

We evaluate four representative VLN backbones on our proposed REVERIE-AIM to validate this gap. As shown in Table~\ref{tab:main_results}, all methods exhibit a severe performance drop between SR and our metrics. Notably, ETPNav-FT scores 34.67\% SR but only 6.32\% ONS@0.1m, meaning over 80\% of ``successes'' fail to actually ground the target. This consistent discrepancy across backbones confirms the last-3-meter gap is a systematic issue of the current evaluation protocol.

\section{The REALM}
\label{sec:method}

\begin{figure}[t]  
\centering
\setlength{\abovecaptionskip}{0pt}
\includegraphics[width=0.975\textwidth]{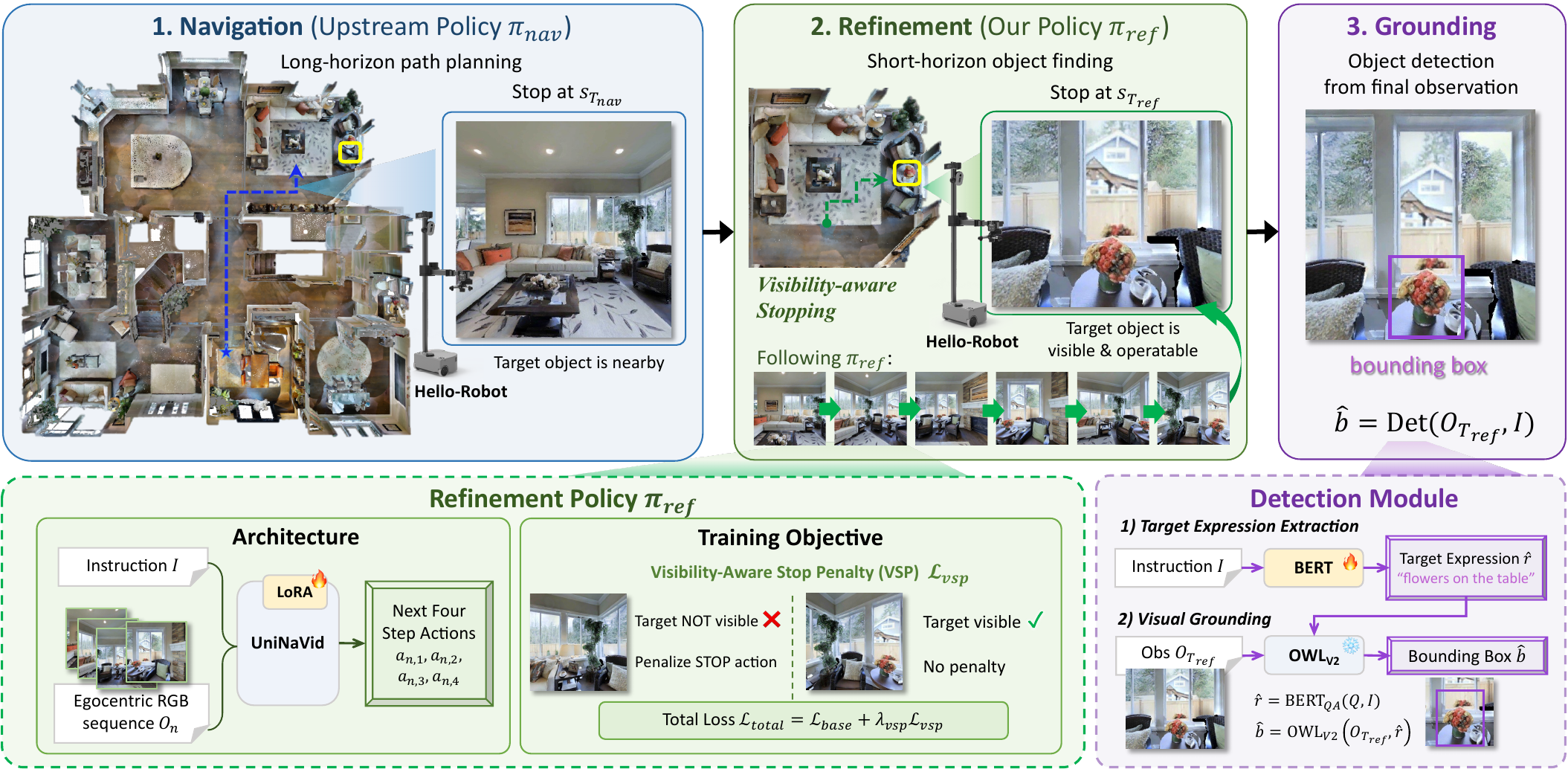}
\caption{\textbf{An overview of the REALM framework.} The pipeline comprises three decoupled stages: \textbf{(i)~Navigation}: $\pi_{\mathrm{nav}}$ follows instruction $\mathcal{I}$ (e.g., \textit{``Go to the living room and water the flowers on the table.''}) and stops at $s_{T_{\text{nav}}}$; \textbf{(ii)~Refinement}: $\pi_{\mathrm{ref}}$ repositions the agent to $s_{T_{\text{ref}}}$ where the target is proximate and visible; \textbf{(iii)~Grounding}: a BERT-based extractor yields a target phrase $\hat{r}$, passed to OWLv2 for bounding box prediction. The refinement module is plug-and-play across arbitrary VLN backbones.}
\label{fig:pipeline} 
\vspace{-5mm}
\end{figure}

We propose \textbf{REALM}, a plug-and-play decoupled refinement module to solve the above last-3-meters grounding problem. The complete pipeline has three stages (Figure~\ref{fig:pipeline}): \textbf{(i)~Navigation}—$\pi_{\mathrm{nav}}$ follows $\mathcal{I}$ and stops at $s_{T_{\text{nav}}}$; \textbf{(ii)~Refinement}—$\pi_{\mathrm{ref}}$ repositions and reorients the agent to a refined state $s_{T_{\text{ref}}}$ where the target is both proximate and visible; \textbf{(iii)~Grounding}—a detection module predicts the bounding box over the final observation. 
To maximize efficiency and flexibility, $\pi_{\mathrm{ref}}$ operates fully independently of the upstream navigation policy $\pi_{\mathrm{nav}}$. This decoupled design avoids performance trade-offs between long-horizon path planning (tens of meters) and short-horizon precision approaching, while ensuring seamless plug-and-play compatibility across diverse VLN backbones (e.g., graph-based, video-based, or zero-shot paradigms).

\subsection{Refinement Policy}
We instantiate the refinement policy $\pi_{\text{ref}}$ with UniNaVid~\cite{uninavid}, a video-based Vision-Language-Action (VLA) foundation model that maps egocentric video streams and language instructions to multi-step action sequences. We choose UniNaVid for two reasons: its video-level temporal encoding naturally suits the sequential observations arising in short-horizon refinement, and its pre-trained visual-linguistic representations provide a strong initialisation that reduces the data requirement for downstream adaptation. At each decision step $t$, the model consumes the egocentric RGB sequence $\mathcal{O}_t$ together with the instruction $\mathcal{I}$, and predicts the next four-step action sequence $a_{t,1}, a_{t,2}, a_{t,3}, a_{t,4}$ over the low-level action space $\mathcal{A} = \{\texttt{forward},\, \texttt{turn\_left},\, \texttt{turn\_right},\, \texttt{stop}\}$.

During training, non-action positions (e.g., instruction and visual tokens) are masked with the ignore index $-100$. Let $\mathcal{M} = \{(b,n) : y_{b,n} \neq -100\}$ denote the set of valid action token positions across a batch of size $B$ with token sequence length $N$. The baseline imitation learning objective minimizing token-level cross-entropy is:
\begin{equation}
    \mathcal{L}_{\mathrm{base}} = \frac{1}{|\mathcal{M}|}\sum_{(b,n)\in\mathcal{M}} \ell_{\mathrm{CE}}\!\left(\hat{\mathbf{z}}_{b,n},\, y_{b,n}\right),
    \label{eq:base_loss}
\end{equation}
where $y_{b,n}$ and $\hat{\mathbf{z}}_{b,n}$ represent the target tokens and predicted logits at batch index $b$ and token position $n$, respectively.

 We apply LoRA~\cite{hu2022lora} to all linear layers of the LLM backbone while keeping the vision encoder frozen. However, directly optimizing $\mathcal{L}_{\mathrm{base}}$ suffers from severe trajectory biases, causing the agent to stop prematurely. We address these failure modes via a visibility-aware stopping strategy.

\vspace{-8pt}
\paragraph{Visibility-Aware Stop Penalty.}

During navigation the agent may issue a \texttt{stop} action before the target is within view, ending the episode prematurely with no chance of recovery.
We penalise this by measuring how strongly the model favours \texttt{stop} and applying a punishment only when the target is absent.

Concretely, let $\hat{\mathbf{z}}_{b} \in \mathbb{R}^{|\mathcal{V}|}$ be the logit vector at the first valid decision step (i.e., $a_{t,1}$ of the four-step sequence $a_{t,1}, a_{t,2}, a_{t,3}, a_{t,4}$ predicted at step $t$) for sample $b$, with $\mathcal{V}$ the action vocabulary, and let $v_b\in\{0,1\}$ indicate target visibility from simulator annotations.
We define the \emph{stop margin} to quantify how strongly the model favors stopping over all alternative low-level actions:
$m_b \;=\; \hat{z}_{b,\,\texttt{stop}} \;-\; \max_{k \neq \texttt{stop}}\, \hat{z}_{b,\,k}\,,
    \label{eq:stop_margin}$
which is positive exactly when \texttt{stop} is the greedy action.
To selectively punish incorrect termination without hindering correct stops, the penalty averages the positive part of this margin over all target-invisible samples:
\begin{equation}
    \mathcal{L}_{\mathrm{vsp}}
    \;=\;
    \frac{1}{N_{\mathrm{inv}}}
    \sum_{b=1}^{B}\,
    \operatorname{ReLU}(m_b)\;\cdot\;\mathbb{I}[v_b = 0]\,,
    \label{eq:vsp_loss}
\end{equation}
where $N_{\mathrm{inv}} = \max\!\bigl(\sum_{b} \mathbb{I}[v_b{=}0],\;1\bigr)$, $\mathbb{I}[\cdot]$ is the indicator function, and the denominator is clamped to at least $1$ for numerical stability. 
This leaves the agent free to stop once the target is perceived, while suppressing premature termination during exploration.

\vspace{-8pt}
\paragraph{Total Training Objective.}
\label{para:total_loss}
The final loss combines imitation and the stop penalty:
\begin{equation}
    \mathcal{L}_{\mathrm{total}}
    \;=\; \mathcal{L}_{\mathrm{base}}
        \;+\; \lambda_{\mathrm{vsp}}\,\mathcal{L}_{\mathrm{vsp}},
    \label{eq:total_loss}
\end{equation}
where $\lambda_{\mathrm{vsp}}$ controls the penalty strength relative to the imitation signal.

\subsection{Target Expression Extraction and Visual Grounding}
\label{sec:target_extraction}

After the refinement policy terminates at $s_{T_{\text{ref}}}$, the agent must localise the target object in its final observation. However, REVERIE-AIM instructions often contain both navigation cues and object descriptions, e.g., \textit{``go to the bathroom and clean the toilet seat''}. Directly using the full instruction as an open-vocabulary detector query may introduce irrelevant spatial or action-related words. We therefore first extract a target phrase from the instruction and then use it for visual grounding.

For target expression extraction, we use GPT-4o-mini to automatically annotate target noun phrases, and fine-tune a BERT-Large~\cite{bert} extractive QA model on 2,048 curated instruction--phrase pairs. At inference time, the BERT model extracts the target span $\hat{r}$ from the instruction 
$\mathcal{I}$, $\hat{r} = \mathrm{BERT}_{QA}(\mathcal{I})$. For visual grounding, the extracted $\hat{r}$ is passed to OWLv2~\cite{owlv2} 
as the text query to predict the final bounding box $\hat{b}$,
$\hat{b} = \mathrm{OWL}_{V2}(\mathcal{O}_{T_{\text{ref}}}, \hat{r})$,
where $O_{T_{\text{ref}}}$ is the agent's final egocentric observation. This design removes irrelevant navigation cues from the grounding query.

\section{The REVERIE-AIM Dataset}
\label{sec:reverie-aim-dataset}
To support the instance-level evaluation metrics introduced in the main content and provide sufficient supervision for both long-horizon navigation and fine-grained target approaching, we construct the REVERIE-AIM dataset by extending the original REVERIE~\cite{qi2020reverie} through three progressive stages. First, we remap the discrete navigation graph into the Habitat continuous simulator to obtain physically plausible trajectories (Section~\ref{sec:con-traj-recon}). Second, we replace the original region-level goal annotations with instance-level object-centric endpoints aligned via ObjectNav sampling (Section~\ref{sec:Object-Centric-Goal-Alignment}). Third, we summarize the dataset statistics and compare against existing benchmarks (Section~\ref{sec:dataset-stats}). We then extract a dedicated short-horizon training set from REVERIE-AIM to provide dense supervision for the final-meters approaching phase (Section~\ref{sec:short-horizon}).

\begin{table}[t]
\centering
\caption{Comparison with existing embodied navigation datasets. Our dataset is the first to combine all four desirable properties.}
\label{tab:datasets}
\resizebox{0.85\columnwidth}{!}{%
\begin{tabular}{lcccc}
\toprule
\textbf{Dataset} & \textbf{Continuous} & \textbf{Language} & \textbf{Instance-level} & \textbf{Object Proximity} \\
 & \textbf{Space} & \textbf{Goal} & \textbf{Target} & \textbf{Endpoint} \\
\midrule
\multicolumn{5}{l}{\textit{Vision-and-Language Navigation (Discrete)}} \\[2pt]
R2R~\cite{anderson2018vision} & \xmark & \cmark & \xmark & \xmark \\
RxR~\cite{anderson2020rxr} & \xmark & \cmark & \xmark & \xmark \\
REVERIE~\cite{qi2020reverie} & \xmark & \cmark & \cmark & \xmark \\
\midrule
\multicolumn{5}{l}{\textit{Vision-and-Language Navigation (Continuous)}} \\[2pt]
R2R-CE~\cite{krantz2020beyond} & \cmark & \cmark & \xmark & \xmark \\
RxR-CE~\cite{anderson2020rxr,krantz2020beyond} & \cmark & \cmark & \xmark & \xmark \\
REVERIE-CE~\cite{reve-ce} & \cmark & \cmark & \cmark & \xmark \\
\midrule
\multicolumn{5}{l}{\textit{Object Navigation}} \\[2pt]
MP3D ObjNav~\cite{batra2020objectnav} & \cmark & \xmark & \xmark & \cmark \\
HM3D ObjNav~\cite{habitatchallenge2022} & \cmark & \xmark & \xmark & \cmark \\
HM3D ImageNav~\cite{hm3d} & \cmark & \xmark & \cmark & \cmark \\
\midrule
\textbf{Ours} & \cmark & \cmark & \cmark & \cmark \\
\bottomrule
\end{tabular}%
}
\end{table}

\subsection{Continuous Trajectory Reconstruction}
\label{sec:con-traj-recon}
The original REVERIE dataset~\cite{qi2020reverie} is constructed upon a discrete navigation graph, where the agent's movement is restricted to transitions between predefined nodes. Such a mechanism fundamentally fails to capture the dynamic complexity of continuous motion and the fine-grained requirements of obstacle avoidance. Inspired by the trajectory transfer strategies adopted in REVERIE-CE~\cite{reve-ce} and Dynam3D~\cite{dynam3d}, we systematically remap the discrete node-based trajectories in REVERIE into the Habitat continuous simulation environment.

Specifically, through trajectory mapping and interpolation, the agent's action space is transformed from discrete node transitions into continuous linear movements and arbitrary-angle rotations. This process ensures that the generated trajectories exhibit both motion smoothness and physical plausibility within a continuous action space.

\subsection{Object-Centric Goal Alignment}
\label{sec:Object-Centric-Goal-Alignment}
The REVERIE dataset defines targets as nearby navigable waypoints rather than the physical boundaries of the instances themselves. Consequently, training on such loosely annotated goals limits agents to region-level arrival instead of precise instance-level localization.

To address this limitation, we integrate the annotation protocol from the Habitat Object Goal Navigation (ObjectNav) task~\cite{onavrim} and propose an object-instance-centric annotation fusion strategy. Capitalizing on the shared Matterport3D scene hierarchy between REVERIE and ObjectNav, we establish a deterministic mapping of instance identifiers across both benchmarks. As a result, the sampling locations generated by ObjectNav for target instances can be seamlessly transferred into our dataset framework without information loss.

Specifically, we replace the original ambiguous nearby-goal annotations with precise sampling points located within the geometric boundary or visually observable region of the target object. This modification elevates the navigation objective from mere \textbf{position reachability} to \textbf{instance-level precise alignment}.

After completing continuous trajectory reconstruction and object-centric goal alignment, the REVERIE-AIM dataset possesses the fundamental structure required for long-horizon navigation training. 

\subsection{Dataset Statistics}
\label{sec:dataset-stats}
The resulting REVERIE-AIM dataset contains 3,691 training trajectories spanning 53 scenes and covering 90 target object categories, as well as 1,344 validation trajectories (\emph{val-unseen}) across 10 unseen scenes with 48 object categories. To contextualize the scalability and structural superiority of REVERIE-AIM within the robot learning community, we present a feature-matrix comparison against existing embodied navigation benchmarks in Table~\ref{tab:datasets}.

Our object-centric goal alignment directly bridges the region-to-object grounding gap. While standard REVERIE-CE endpoints terminate at a loose, region-level distance, REVERIE-AIM reduces the valid average minimum geodesic distance to the target from $1.32$ m to just $0.34$ m. This finer-grained spatial resolution enables more precise evaluation of both navigation and instance-level grounding.

\subsection{Short-Horizon Training Data Generation}
\label{sec:short-horizon}
We further construct a dedicated fine-grained navigation dataset based on REVERIE-AIM. The primary objective of this stage is to generate a large number of short-horizon video clips containing \emph{instruction--visual observation--action sequence} tuples, thereby enhancing the agent's decision-making capability during the target interaction stage.

\paragraph{Task-specific Initial Sampling}
To simulate the visual state of an agent approaching the final stage of long-horizon navigation, we select the second-to-last node of the reference path as the anchor point for initialization. To further improve the model's tolerance to localization errors and enhance generalization, we augment the initial states through the following strategies:

\noindent\textbf{Spatial Perturbation.}  
We perturb the anchor position using a truncated Gaussian distribution ($\sigma = 1.5$, truncated within $[-5, 5]$ meter) to generate spatially diverse starting states around the target object.

\noindent\textbf{Viewpoint Augmentation.}  
We randomly sample the initial heading angle within $[-\pi, \pi]$ to simulate diverse target-approaching viewpoints.

\paragraph{Clip and Action Sequence Extraction}

For each valid sampled starting point, we compute the shortest path to the target object's surrounding sampling location, determined by the object-centric endpoint described in Section~\ref{sec:Object-Centric-Goal-Alignment}. We then extract the corresponding visual observation sequences (video clips) together with their continuous action labels. The action labels are a set of low-level actions:  $\{\texttt{forward},\, \texttt{turn\_left},\, \texttt{turn\_right},\, \texttt{stop}\}$.
In total, the resulting dataset contains approximately 180,000 fine-grained navigation samples from the training split. These samples provide dense supervision for learning precise short-horizon navigation behaviors, particularly during the final target-approaching stage.
\section{Experiments}

\vspace{-5pt}

\subsection{Experimental Setup}

\paragraph{Upstream VLN Baselines}
We evaluate our plug-and-play module across four diverse VLN-CE baselines to verify its universality: (1) ETPNav-ZS~\cite{an2024etpnav}, a zero-shot topological planning policy, testing refinement under zero exposure to object-referring instructions; (2) ETPNav-FT~\cite{an2024etpnav}, a fine-tuned variant of ETPNav, evaluating benefits when the upstream policy is already task-calibrated; (3) UniNaVid-ZS~\cite{uninavid}, a video-based VLA baseline that shares the same base architecture as our module, isolating the gains of our LoRA adaptation; and (4) Smartway~\cite{smartway}, a training-free paradigm using large VLMs. These backbones comprehensively cover zero-shot/fine-tuned, graph/video-based, and training-free navigation paradigms.

\vspace{-10pt}
\paragraph{Downstream Variants}
To verify the components and optimization of our refinement module, we compare six downstream variants: 
(1)w/o Refinement, which directly uses the upstream policy's raw endpoints;
(2) Random Walk, which performs random movements at the destination; 
(3) Heuristic CLIP-Search, which spins $360^\circ$ and moves toward the direction with the highest CLIP matching score; 
(4) Vanilla UniNaVid, the training-free base architecture executing raw upstream endpoints directly; 
(5) Ours w/o VSP, our model optimized on our data using standard cross-entropy loss; and 
(6) Ours (Full), our complete module optimized on our data by the full tailored loss function.

\vspace{-8pt}
\paragraph{Standard VLN-CE Metrics}
We report standard VLN-CE metrics (TL, NE, OSR, SR, SPL)~\cite{hong2022bridging} for upstream reference. our instance-level metrics (ONS, GS, OracleGS) are the primary evaluation targets.

\begin{table*}[h]
\centering
\caption{
  Plug-in fine-grained navigation module performance on REVERIE-AIM (val-unseen).
  Standard VLN metrics are fixed per backbone and shown in parentheses beside each backbone name.
  \textbf{Bold} = best in group; \underline{underline} = second best.
}
\label{tab:main_results}
\setlength{\tabcolsep}{9pt}      
\renewcommand{\arraystretch}{0.7} 
\setlength{\aboverulesep}{0.25ex}
\setlength{\belowrulesep}{0.25ex}
\setlength{\extrarowheight}{0pt}
\begin{adjustbox}{max width=\textwidth}
\begin{tabular}{ll cccc}
\toprule
\textbf{Backbone}
  & \textbf{Move Policy}
  & ONS@0.1m$\uparrow$
  & ONS@0.5m$\uparrow$
  & GS$\uparrow$
  & OracleGS$\uparrow$ \\
\midrule
\multicolumn{6}{l}{\textit{ETPNav-ZS}
  \;\textcolor{gray}{\small TL\,=\,11.85,\; NE$\downarrow$\,=\,7.26,\; OSR\,=\,35.86,\; SPL\,=\,23.39,\; SR\,=\,28.94}} \\
\midrule
\multirow{6}{*}{\rotatebox[origin=c]{90}{\small ETPNav-ZS}}
  & --- (no move)
    & 7.07  & 11.98 & 3.05 & 6.99 \\
  & Random
    & 5.21  & 10.42 & 1.93 & 4.61 \\
  & Heuristic
    & 7.51  & 12.57 & 4.02 & 10.19 \\
  & Vanilla UniNavid
    & 11.16 & 17.78 & 4.54 & 11.90 \\
  & Ours w/o VSP
    & \underline{13.47} & \underline{20.24} & \underline{5.28} & \underline{15.55} \\
  & \textbf{Ours (Full)}
    & \textbf{14.66} & \textbf{21.13} & \textbf{6.10} & \textbf{16.52} \\
\midrule
\multicolumn{6}{l}{\textit{ETPNav-FT}
  \;\textcolor{gray}{\small TL\,=\,13.86,\; NE$\downarrow$\,=\,6.71,\; OSR\,=\,40.63,\; SPL\,=\,27.68,\; SR\,=\,34.67}} \\
\midrule
\multirow{6}{*}{\rotatebox[origin=c]{90}{\small ETPNav-FT}}
  & --- (no move)
    & 6.32  & 12.20 & 4.46  & 11.31 \\
  & Random
    & 5.65  & 12.95 & 2.75  & 8.78 \\
  & Heuristic
    & 6.77  & 13.24 & 5.88  & 12.05 \\
  & Vanilla UniNavid
    & 13.39 & 19.12 & 4.91  & 12.57 \\
  & Ours w/o VSP
    & \underline{16.82} & \underline{25.00} & \underline{8.04} & \underline{19.79} \\
  & \textbf{Ours (Full)}
    & \textbf{17.19} & \textbf{25.89} & \textbf{8.48} & \textbf{20.24} \\
\midrule
\multicolumn{6}{l}{\textit{UniNavid-ZS}
  \;\textcolor{gray}{\small TL\,=\,16.07,\; NE$\downarrow$\,=\,8.69,\; OSR\,=\,34.40,\; SPL\,=\,17.40,\; SR\,=\,25.10}} \\
\midrule
\multirow{6}{*}{\rotatebox[origin=c]{90}{\small UniNavid-ZS}}
  & --- (no move)
    & 13.61 & 18.08 & 5.65  & 12.20 \\
  & Random
    & 9.75  & 16.89 & 2.83  & 8.78 \\
  & Heuristic
    & 12.35 & 18.15 & 5.13  & 10.42 \\
  & Vanilla UniNavid
    & 10.42 & 16.07 & 4.84  & 11.01 \\
  & Ours w/o VSP
    & \underline{12.43} & \underline{18.38} & \underline{5.95} & \underline{14.29} \\
  & \textbf{Ours (Full)}
    & \textbf{14.96} & \textbf{20.68} & \textbf{6.85} & \textbf{16.37} \\
\midrule
\multicolumn{6}{l}{\textit{Smartway}
  \;\textcolor{gray}{\small TL\,=\,20.00,\; NE$\downarrow$\,=\,8.79,\; OSR\,=\,25.37,\; SPL\,=\,4.81,\; SR\,=\,8.21}} \\
\midrule
\multirow{6}{*}{\rotatebox[origin=c]{90}{\small Smartway}}
  & --- (no move)
    & 7.81  & 11.83 & 1.49  & 4.69 \\
  & Random
    & 5.21  & 10.42 & 1.34  & 4.61 \\
  & Heuristic
    & 7.22  & 11.98 & 2.90  & 8.04 \\
  & Vanilla UniNavid
    & 9.30  & 14.81 & 3.50  & 9.45 \\
  & Ours w/o VSP 
    & \underline{12.95} & \underline{17.49} & \underline{4.69} & \underline{14.66} \\
  & \textbf{Ours (Full)}
    & \textbf{13.99} & \textbf{19.72} & \textbf{5.65} & \textbf{15.92} \\
\bottomrule
\multicolumn{2}{l}{\textit{Human Eval}}
  & 50.40 & 74.71 & 38.24 & 80.38 \\
\bottomrule
\end{tabular}
\end{adjustbox}
\vspace{-5mm}

\end{table*}

\subsection{Results in Simulator}
As shown in Table~\ref{tab:main_results}, our full module (\textbf{Ours (Full)}) consistently and significantly outperforms all alternative move policies across all four backbones. For instance, on ETPNav-FT, it boosts ONS@0.1m from $6.32\%$ (no move) to $\mathbf{17.19\%}$ and GS from $4.46\%$ to $\mathbf{8.48\%}$, demonstrating its superior capability in fine-grained object localization. The ablation variant (Ours w/o VSP) incurs consistent performance drops across all settings (e.g., ONS@0.1m falls by $2.53\%$ on UniNavid-ZS), validating the necessity of our proposed Visibility-Aware Stopping loss. Nevertheless, a substantial performance gap remains compared to Human Eval (e.g., $25.89\%$ vs. $74.71\%$ on ONS@0.5m), highlighting that this benchmark leaves massive room for future improvement.

\subsection{Results in Real-world Environments}
To verify REALM in real-world performance, we deploy both the vanilla UniNavid and our enhanced pipeline on a Hello Robot Stretch platform equipped with an Intel RealSense D435if RGB-D camera. All real-world experiments are conducted on a workstation with the NVIDIA RTX 4090 GPU. Since real-world environments lack predefined navigable viewpoints, we define ONS@0.5m as the Euclidean distance between the robot and the nearest navigable position to the instance. We collect 12 diverse episodes across multiple rooms, including a robot lab, the school's lobby and foyer. Since UniNavid already performs well in simulation without our module, we compare the vanilla UniNavid against UniNavid as Upstream module augmented with our method.

\begin{wraptable}{r}{0.60\textwidth}
\vspace{-18pt}
\centering
\caption{
  Comparison on Real-world robot evaluation.
}
\label{tab:real_robot}
\setlength{\tabcolsep}{6pt}
\begin{tabular}{l ccc}
\toprule
\textbf{Method}
  & ONS@0.5m$\uparrow$ & GS$\uparrow$ & OracleGS$\uparrow$   \\
\midrule
UniNavid
  & 8.33 & 8.33  & 16.67  \\
UniNavid + Ours
  & \textbf{33.33}& \textbf{25.00} & \textbf{25.00}  \\
\bottomrule
\end{tabular}
\end{wraptable}

As shown in Table~\ref{tab:real_robot}, the vanilla UniNavid policy struggles severely in the physical environment, yielding an 8.33\% SR due to accumulated motor noise and its limited exposure to REVERIE-style instructions during training, alongside a low OracleGS of 16.67\% and GS of 8.33\%. In contrast, our plug-and-play refinement module (UniNavid + Ours) brings substantial improvements across all metrics, escalating the SR to 33.33\%. Crucially, it boosts both OracleGS and GS to 25.00\%. This clear performance gain demonstrates that our visibility-aware strategy successfully corrects upstream drift and aligns the robot's heading, effectively bridging the Last-3-Meter Grounding Gap to provide a reliable perception interface for physical interaction.

\section{Conclusion and Limitations}

We identify the Last-3-Meter Grounding Gap in continuous VLN and introduce REVERIE-AIM, three instance-centric metrics, and REALM to bridge region-level arrival and object-level grounding. Although REALM consistently improves diverse VLN backbones, its absolute performance remains far below the human upper bound. This suggests that final-stage object-centric navigation requires more than short-horizon motion refinement: agents must actively select informative viewpoints, reason about target visibility, and make reliable stopping decisions under partial observability. These challenges remain open for future work.



\bibliography{clean_ref}  

\end{document}